\documentclass[letterpaper, 10 pt, conference]{ieeeconf}  

\IEEEoverridecommandlockouts                              

\usepackage{graphics} 
\usepackage{epsfig} 
\usepackage{mathptmx} 
\usepackage{times} 
\usepackage{amsmath} 
\usepackage{amssymb}  
\usepackage{xcolor}
\usepackage{multirow}
\usepackage{booktabs}
\usepackage{array}
\usepackage{float}
\usepackage{caption}
\usepackage{tabularx}
\usepackage{adjustbox}
\usepackage{threeparttable}
\usepackage{makecell} 
\usepackage{cite}
\usepackage{xcolor}
\usepackage{pifont}

\newcolumntype{C}{>{\centering\arraybackslash}X}

\usepackage{hyperref}
\definecolor{urlblue}{RGB}{0,0,238}
\definecolor{linkpink}{HTML}{F03090} 
\usepackage{makecell}

\title{\LARGE \bf
EgoTraj-Bench: Towards Robust Trajectory Prediction Under Ego-view Noisy Observations
}

\author{
Jiayi Liu$^{1}$, 
Jiaming Zhou$^{1}$, 
Ke Ye$^{1}$, 
Kun-Yu Lin$^{2}$, 
Allan Wang$^{3}$, 
and Junwei Liang$^{1,4}$\textsuperscript{*}
\thanks{$^{1}$The Hong Kong University of Science and Technology (Guangzhou). 
\texttt{Jiayi.LIU@connect.hkust-gz.edu.cn}, 
\texttt{junweiliang@hkust-gz.edu.cn} * Corresponding author.}
\thanks{$^{2}$The University of Hong Kong.}
\thanks{$^{3}$Miraikan -- The National Museum of Emerging Science and Innovation.}
\thanks{$^{4}$The Hong Kong University of Science and Technology.}
\thanks{This work was supported by the National Natural Science Foundation of China (No. 62306257).}
}

\begin{document}

\maketitle
\thispagestyle{empty}
\pagestyle{empty}

\begin{abstract}

Reliable trajectory prediction from an ego-centric perspective is crucial for robotic navigation in human-centric environments. However, existing methods typically assume noiseless observation histories, failing to account for the perceptual artifacts inherent in first-person vision, such as occlusions, ID switches, and tracking drift.
This discrepancy between training assumptions and deployment reality severely limits model robustness. 
To bridge this gap, we introduce EgoTraj-Bench, built upon TBD dataset, which is the first real-world benchmark that aligns noisy, first-person visual histories with clean, bird's-eye-view future trajectories, enabling robust learning under realistic perceptual constraints.
Building on this benchmark, we propose \textbf{BiFlow}, a dual-stream flow matching model that concurrently denoises historical observations and forecasts future motion.
To better model agent intent, BiFlow incorporates our \textbf{EgoAnchor} mechanism, which conditions the prediction decoder on distilled historical features via feature modulation. 
Extensive experiments show that BiFlow achieves state-of-the-art performance, reducing minADE and minFDE by 10–15\% on average and demonstrating superior robustness. 
We anticipate that our benchmark and model will provide a critical foundation for robust real-world ego-centric trajectory prediction. 
The benchmark library is available at: {\color{linkpink}\href{https://github.com/zoeyliu1999/EgoTraj-Bench}{\texttt{https://github.com/zoeyliu1999/EgoTraj-Bench}}}.


\end{abstract}

\section{Introduction}

\begin{figure}[htbp]
    \centering
    \makebox[\linewidth][c]{\includegraphics[width=0.9\linewidth,trim={0cm 0cm 0cm 0cm}, clip]{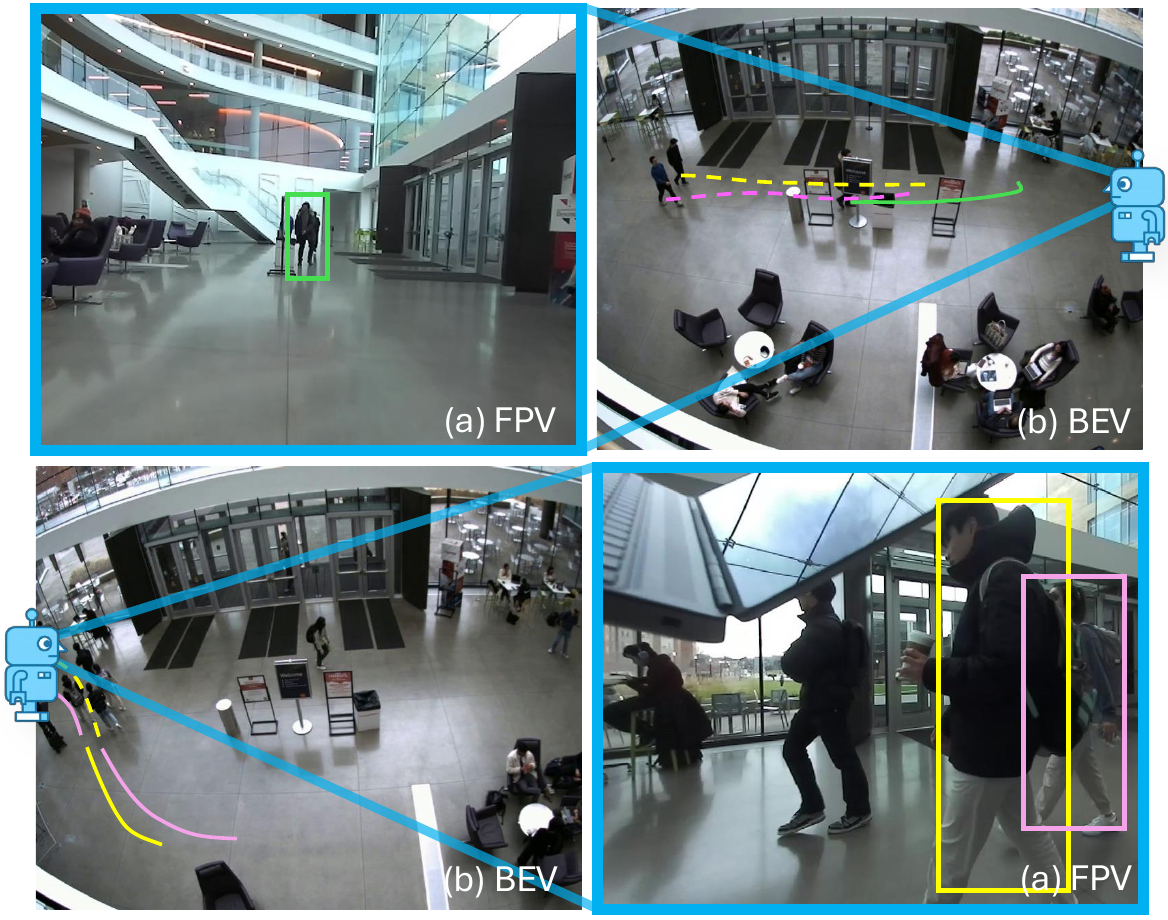}}
    \caption{Illustration of key challenges under ego-view observations. \textbf{Top row: Occlusion.} In the first-person view (a), only one pedestrian (green box) is visible; the corresponding BEV (b) shows two additional agents (pink and yellow) behind the green pedestrian. Dashed lines indicate trajectories visible in BEV but not in FPV. \textbf{Bottom row: ID Switch and Perspective Distortion.} Two pedestrians (yellow and pink) cross paths, causing an ID swap in the FPV tracking output (a). Additionally, individuals near the image corners suffer significant perspective distortion, making accurate localization challenging.}
    \label{fig:intuition}
    \vspace{-1em}
\end{figure}

Pedestrian trajectory prediction\cite{alahi2016social, gupta2018social}, aiming to estimate the multimodal future paths of agents in dynamic environments, serves as a foundation for safe, socially compliant motion planning in autonomous systems such as mobile robots, intelligent prosthetics, and service vehicles\cite{sun20183dof, tonoki2017model, li2020socially, chen2023future, 10341464}. 
Although extensively studied, most existing methods are developed and evaluated under idealized bird’s-eye view (BEV) settings with globally consistent observations and flawless agent tracking\cite{gupta2018social}. However, these conditions rarely hold in real-world deployment. Autonomous agents, such as mobile robots, typically perceive the environment through front-facing cameras, where observations are inherently incomplete and noisy as illustrated in Fig.~\ref{fig:intuition}: pedestrians may be occluded, enter or exit the field of view (FOV), or suffer from physical sensing errors such as perspective distortion. These imperfections substantially violate the idealized historical assumptions in BEV settings. Therefore, trajectory prediction under ego-centric (first-person view, FPV) noisy observations is essential for enabling robust deployment in real-world scenarios.
\vspace{-0.5em}

Some prior work~\cite{yagi2018future,chen2023future, rasouli2024novel} predicts trajectories from ego-centric input, typically predicting future positions in image space, e.g., bounding boxes or keypoints. Despite operating from an ego-centric view, these methods lack spatial reasoning in real-world space, assume idealized tracking results in image space, and thus leave the modeling of fine-grained interactions unresolved in real-world trajectory prediction. In contrast, methods\cite{salzmann2020trajectron++, wang2022stepwise} that predict trajectories in global metric spaces (e.g., world coordinates) enable precise spatial reasoning about proximity, collision risk, and social norms, and we therefore focus on the latter paradigm. 

In addition, some studies~\cite{bi2020can, stoler2023t2fpv} simulate ego-centric conditions by rendering BEV data into synthetic views using simulators. While this approximates the visual input of moving entities, the rule-based agent motion and simplified rendering in the simulator fail to capture the intricate and subtle motion patterns and visual nuances present in authentic scenes. Moreover, the utilized BEV data\cite{lerner2007crowds, pellegrini2010improving} are collected in open and uncluttered environments such as streets with few static obstacles, resulting in overly clean inputs that do not reflect the perception challenges of dense, interactive environments. These limitations highlight the critical need for a real-world benchmark for robust trajectory prediction with ego-centric noisy observations.

To this end, we introduce EgoTraj-Bench, the first real-world benchmark for trajectory prediction under ego-centric noise. Built upon the TBD dataset\cite{wang2024tbd}, EgoTraj-Bench first derives historical trajectories with noise from real ego-view videos, capturing deployment-realistic imperfections such as occlusions, mis-tracked IDs, FOV truncations, and perspective distortions. Furthermore, the observed ego-centric trajectory with noise is projected into world coordinates and paired with the corresponding clean, human-verified future trajectory from the BEV view, ensuring metric-consistent supervision while preserving the realism of ego-centric input conditions. This practice can transfer the disturbance from the ego-view noise to the widely used BEV-based trajectory prediction framework, thereby providing a fairer and trustworthy platform for systematically evaluating existing BEV-based trajectory prediction methods. The benchmarking results show that state-of-the-art BEV-based models suffer significant performance degradation when their input of historical observations is disturbed by the ego-view noises, underscoring the need for new frameworks for robust trajectory prediction under real-world ego-view perturbation.


To address this problem, we propose BiFlow, a novel noise-resistant dual-stream flow matching model as an example solution for our benchmark. BiFlow jointly recovers the observed noisy historical trajectories and predicts future trajectories. By jointly learning latent features across the two tasks, the model implicitly leverages denoised historical semantics to guide future trajectory predictions, improving robustness while maintaining parameter efficiency. In addition, we introduce EgoAnchor, a mechanism to distill compact, ego-centric tokens from agent- and scene-level histories. These intent-aware representations, extracted via attention mechanism during history reconstruction, are injected into the decoder via feature-wise affine modulation, providing a robust intent prior to stabilize prediction under partial or corrupted input. 

The main contributions of our work are: 1) We introduce EgoTraj-Bench, the first real-world benchmark for trajectory prediction under deployment-realistic conditions, enabling rigorous evaluation of models under authentic ego-centric noisy perturbations; 2) We propose a novel dual-stream flow matching framework with a distillation mechanism, which jointly recovers noisy historical observations and predicts future trajectories, aiming to leverage clean historical semantics to facilitate and stabilize future forecasting; 3) Our experiments demonstrate the significant impact of ego-view noise on existing models and the robustness of our proposed approach, which outperforms baselines by over 10\% in minADE and 13\% in minFDE averaged over datasets, highlighting the importance of noise-aware modeling and providing valuable insights for future research in ego-view realistic trajectory prediction. 

\begin{figure*}
    \centering    \includegraphics[width=1\linewidth, trim={0cm 0cm 0cm 0cm}, clip]{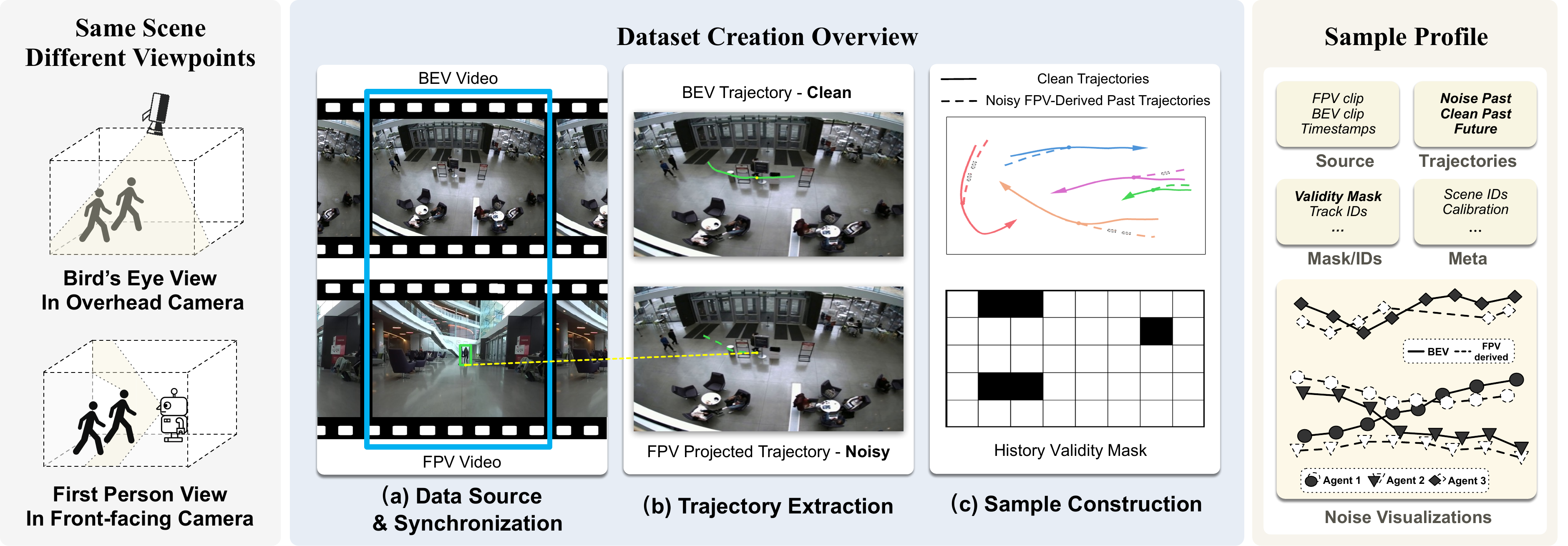}
    \caption{EgoTraj-Bench Overview: \textbf{Left} contrasts the same scene in BEV vs FPV, where limited visibility and viewpoint changes introduce noisy histories. \textbf{Mid} shows sample construction: synchronized BEV–FPV clips provide noisy FPV-derived past trajectories as inputs and clean metric BEV trajectories as supervision, with a history validity mask. \textbf{Right} illustrates what the benchmark records and the key noise patterns that drive robustness gaps beyond idealized BEV settings.}
    \label{fig:benchmark}
    \vspace{-1em}
\end{figure*}

\section{Related Work}

\subsection{Pedestrian Trajectory Prediction}

Various models have been proposed for pedestrian trajectory prediction. Social-LSTM\cite{alahi2016social} pioneers agent interaction modeling via pooled LSTM states. Social GAN\cite{gupta2018social} and AC-VRNN\cite{bertugli2021ac} introduce generative frameworks to capture multimodal futures. More recently, Transformer-based architectures like TUTR\cite{shi2023trajectory} leverage self-attention for long-range spatiotemporal reasoning. 
Denoising models, particularly diffusion models and flow-based generative models\cite{ho2020denoising}, further improve the modeling of complex multimodal trajectory distributions via iterative refinement. MID\cite{gu2022stochastic} first applies diffusion to predict future trajectory, while Leapfrog\cite{mao2023leapfrog} introduces a deterministic latent initializer to speed up sampling. However, both sample futures independently per agent, leading to spatially redundant trajectories. MoFlow\cite{fu2025moflow} uses flow matching (FM) to jointly model futures conditioned on past motion, promoting learning of diverse future trajectories. These advances motivate adopting FM to generate future trajectories from partial and noisy ego-centric inputs, leveraging its ability to condition on corrupted histories and support diverse predictions.

Beyond architecture, several methods incorporate auxiliary inputs, e.g., goal estimates\cite{mangalam2021goals, wang2022stepwise} or scene graphs\cite{salzmann2020trajectron++}, to enrich contextual cues. However, these typically assume clean, globally observed trajectories, an assumption violated in real-world ego-centric settings. While some studies attempt to enhance robustness via random history masking\cite{liu2019naomi, qi2020imitative, chib2024ms}, such artificial missingness fails to reflect the complex and structured nature of real ego-centric perception. This gap motivates the need for trajectory prediction approaches under realistic ego-centric perturbations.


\subsection{Trajectory Prediction under Ego-centric Noise}
Recent efforts address ego-centric prediction via FPV or BEV paradigms. FPV-based methods, such as \cite{yagi2018future} and \cite{rasouli2024novel}, predict pedestrian locations (e.g., bounding boxes or keypoints) in image space. While effective for visual tracking or localization, these rely on idealized image-space observations and unscaled pixel coordinates, limiting the ability to support metric-space reasoning, such as real-world proximity or collision risk. 

BEV-based approaches project perturbed ego-centric trajectories to world coordinates, enabling prediction within a shared metric framework (e.g., in meters), where spatial relationships can be precisely computed. Fvtraj\cite{bi2020can} and T2FPV\cite{stoler2023t2fpv} generate synthetic FPV videos by rendering BEV trajectories in simulators. While enabling controlled study of perception noise, their rule-based agent behaviors and simplified visual rendering fail to capture subtle motion patterns and realistic visual nuances, consequently limiting the model's realism and generalizability. Moreover, underlying BEV datasets (i.e, ETH-UCY\cite{lerner2007crowds, pellegrini2010improving}) were collected in open, sparse environments with minimal obstacles, yielding rendered FPV sequences failing to reflect the dense, dynamic scenes typical of real-world deployment. While several real-world ego-centric datasets exist\cite{ caesar2020nuscenes, martin2021jrdb, grauman2022ego4d}, most focus on past or present state estimation without providing future ground truth in world coordinates, or involve agents (e.g., vehicles) with motion patterns fundamentally different from pedestrian-scale robots. 

Methodologically, T2FPV’s CoFE module denoises historical perturbations using clean history as supervision\cite{stoler2023t2fpv}, but only corrects missing positions when constructing the input of prediction model. This hybrid representation, combining uncorrected observed positions with corrected ones, risks spatial-temporal discontinuities at segment boundaries, potentially causing artificial motion jumps during prediction. 
To this end, this motivates a unified framework that avoids such patchwork reconstruction, together with a real-world benchmark capturing both noisy perception inputs and metric-accurate ground truth.

\section{EgoTraj-Bench: Ego-view Trajectory Prediction Benchmark}

\begin{table*}
\centering
\begin{threeparttable}
\caption{Dataset Statistics Comparison.}
\label{tab:benchmark}
\begin{tabular*}{\textwidth}{@{\extracolsep{\fill}} ll|cccccccc@{}}
\toprule
\multirow{2}{*}{Dataset} & \multirow{2}{*}{Fold} & Duration & Label Freq  &\multirow{2}{*}{\# Traj} & \multicolumn{2}{c}{Ego-noise Involved} & \multirow{2}{*}{Agent Scale} & \multirow{2}{*}{FPV Noisy Rate$^{\dagger}$} & History MSE$^{*}$  \\
\cmidrule(lr){6-7}
 &  & (min) & (Hz) &  & Perceptual & Real-world Physical &  &  & (m) \\
\addlinespace[1ex]
\cline{1-10} 
\addlinespace[-0.5ex] 
\cline{1-10} 
\addlinespace[1ex]
WildTrack\cite{chavdarova2018wildtrack}  & --  & 3 & 2 & 313  & \ding{55}  & \ding{55}  & pedestrian & -- & -- \\
TH{\"o}R\cite{rudenko2020thor}  & --  & 60 & 100 & 600  & \ding{52}  & \ding{55}  & pedestrian & -- & -- \\
nuScenes\cite{caesar2020nuscenes}  & --  & 330 & 12 & -  & \ding{52}  & \ding{52}  & \ding{55} vehicle & -- & -- \\
\multirow{5}{*}{\makecell[l]{T2FPV--ETH\\\cite{stoler2023t2fpv}}}
 & ETH   & 9   & 15   & 181    & \ding{52} & \ding{55} & pedestrian & 0.44 & 5.66 \\ 
 & Hotel & 13  & 15   & 1,053  & \ding{52} & \ding{55}  & pedestrian & 0.51 & 4.55 \\
 & Zara1 & 6   & 2.5  & 5,939  & \ding{52} & \ding{55} & pedestrian & 0.28 & 1.23 \\ 
 & Zara2 & 7   & 2.5  & 17,608 & \ding{52} & \ding{55} & pedestrian & 0.32 & 2.50 \\
 & Univ  & 3.5 & 2.5  & 24,334 & \ding{52} & \ding{55} & pedestrian & 0.45 & 2.47 \\
\addlinespace[1ex]
\cline{1-10} 
\addlinespace[-0.5ex] 
\cline{1-10} 
\addlinespace[1ex]
EgoTraj-TBD & -- & 210 & 30 & 36,947 & \ding{52} & \ding{52} & pedestrian & 0.37 & 0.66 \\
\bottomrule
\end{tabular*}
\begin{tablenotes}
\small
\item[$\dagger, *$] \textbf{FPV noisy rate} indicates the average probability of being marked as invalid. \textbf{History MSE} indicates the distance between estimated and ground-truth history. Further details, such as density statistics, are available in TBD\cite{wang2024tbd}.
\end{tablenotes}
\end{threeparttable}
\vspace{-1em}
\end{table*}

To bridge the gap between idealized BEV-based trajectory prediction and real-world deployment under ego-view perception noise, we introduce EgoTraj-Bench, a novel real-world benchmark that transfers realistic ego-view induced noise into BEV coordinate space, enabling fair evaluation of existing BEV-based models and fine-grained spatial reasoning under deployment-realistic conditions.

\subsection{Real-world Dataset Creation}
\label{sec:data-creation}

We construct the core of EgoTraj-Bench by deriving noisy trajectories from real-world first-person video as shown in Fig.~\ref{fig:benchmark}. Rather than simulating perception artifacts as in \cite{stoler2023t2fpv}, we capture authentic perception artifacts, e.g, occlusions, identity switches, and ego-motion drift, arising from dynamic human-robot interactions and physical sensor limitations in unstructured environments.

The generated pixel-space trajectories from authentic videos are projected into the global BEV coordinate system using calibrated camera intrinsics and time-synchronized robot ego-motion (the information recorded during data collection), ensuring metric consistency with BEV-based prediction models. Each ego-view-derived noisy history trajectory in BEV space is then temporally aligned with its corresponding clean past and future trajectory extracted from overhead cameras. This paired structure, i.e, noisy history as input, clean past and future as supervision, enables rigorous and fair evaluation of trajectory prediction robustness under realistic ego-view noise, while preserving the spatial reasoning supporting navigation in crowded scenes.

In the following sections, we detail the full pipeline across three stages: Data Source and Synchronization, Trajectory Extraction, and Sample Construction.

\noindent\textbf{- Data Source and Synchronization.} To establish a foundation for injecting and evaluating real-world ego-view noise, we build upon the publicly available TBD dataset\cite{wang2024tbd}, which uniquely provides synchronized BEV videos from overhead cameras and ego-view videos from mobile robots. While TH\"or\cite{rudenko2020thor} also offers dual-view data, it was collected in a controlled laboratory setting, resulting in less natural human behavior and reduced environmental diversity. We select segments where the robot is actively moving and collecting pedestrian data, and for each ego-view clip, extract its temporally aligned BEV counterpart. This synchronization allows every trajectory extracted from ego-view video, inherently noisy and perspective-distorted, to be geometrically projected into the shared world coordinate system, where they can be direct spatially aligned with clean ground-truth trajectories for fair and consistent evaluation. 

\noindent\textbf{- Trajectory Extraction.} To ensure perception noise reflects real deployment conditions, we extract pedestrian trajectories from raw FPV video using YOLOv8\cite{yolov8_ultralytics} for detection, selected for its robustness in crowded scenes, and BotSort\cite{aharon2022bot} for tracking, which fuses motion and appearance cues to mitigate ID switches. Visibility is quantified per frame via segmentation masks from YOLOv8-seg. All hyperparameters are tuned for different scenes. The final 2D bounding box bottom centers are back-projected to the ground plane using calibrated intrinsics and synchronized ego-motion from the TBD dataset, yielding BEV-space trajectories that retain realistic noise such as occlusion, ID instability, and localization error. For the supervision of clean past and future trajectories, we use the BEV annotations provided in the TBD dataset, benefiting from occlusion-resilient multi-view coverage and semi-automated human verification.

\noindent\textbf{- Sample Construction.} To enable supervised learning under ego-view noise, we align each noisy ego-view derived history with its corresponding clean past and future trajectory from BEV annotations. Due to unaligned track IDs from independent FPV and BEV pipelines, input and ground-truth tracks are associated using Hungarian matching as in \cite{wang2024tbd, weng2022whose, stoler2023t2fpv}. Instead of relying solely on mean squared error (MSE) in absolute position, we compute weighted MSEs incorporating location, velocity, and acceleration to enhance matching robustness under noise. Following common practice\cite{lerner2007crowds, pellegrini2010improving, robicquet2017learning}, we adopt an 8-second sliding window for each sample: 8 frames (3.2 s) for observation and subsequent 12 frames (4.8 s) for prediction. Only samples with at least three valid observation frames are retained, where validity is determined by visibility (more than 100 pixels in the segmentation mask) and motion plausibility (estimated frame speed less than 2 m/s). Missing observations within valid samples are marked and can be filled via techniques such as linear interpolation to ensure fixed-length inputs, while the robot’s trajectory is included to model agent-robot interaction. The final dataset contains 36,947 aligned pairs, each linking noisy BEV-aligned histories to clean BEV past and future.

\subsection{Statistics and Analysis}

As summarized in Table~\ref{tab:benchmark}, EgoTraj-Bench provides 210 minutes of real-world recordings at 30 Hz, offering extended interaction diversity and fine temporal resolution for ego-centric modeling. While the total number of detected trajectories is comparable to synthetic benchmarks T2FPV-ETH, the generation strategy differs fundamentally: T2FPV-ETH treats every agent as a virtual ego-observer, artificially multiplying samples per scene, whereas EgoTraj-Bench preserves natural perceptual bias by including only one true mobile observer per scene. 

EgoTraj-TBD covers two key challenge types: (1) inherent perceptual artifacts (e.g., occlusion, FOV truncation, ID switching); and (2) physical sensor artifacts (e.g., lens distortion, projection error, ego-motion drift), important to real-world systems.
EgoTraj-Bench also focuses on pedestrian-scale agents in human-centric environments, whose surroundings and motion patterns differ fundamentally from vehicle-centric benchmarks such as nuScenes~\cite{caesar2020nuscenes}. 
As reported in Table~\ref{tab:benchmark}, the dataset features a moderate noise rate and lower historical alignment error, which could be explained by higher-fidelity ground truth and a more noise-aware processing pipeline as mentioned in Sec.~\ref{sec:data-creation}. During learning, the dataset is chronologically split into train, validation, and test sets (70\%-10\%-20\%) to ensure temporal coherence and avoid data leakage.

\begin{figure*}
    \centering
    \includegraphics[width=0.98\linewidth]{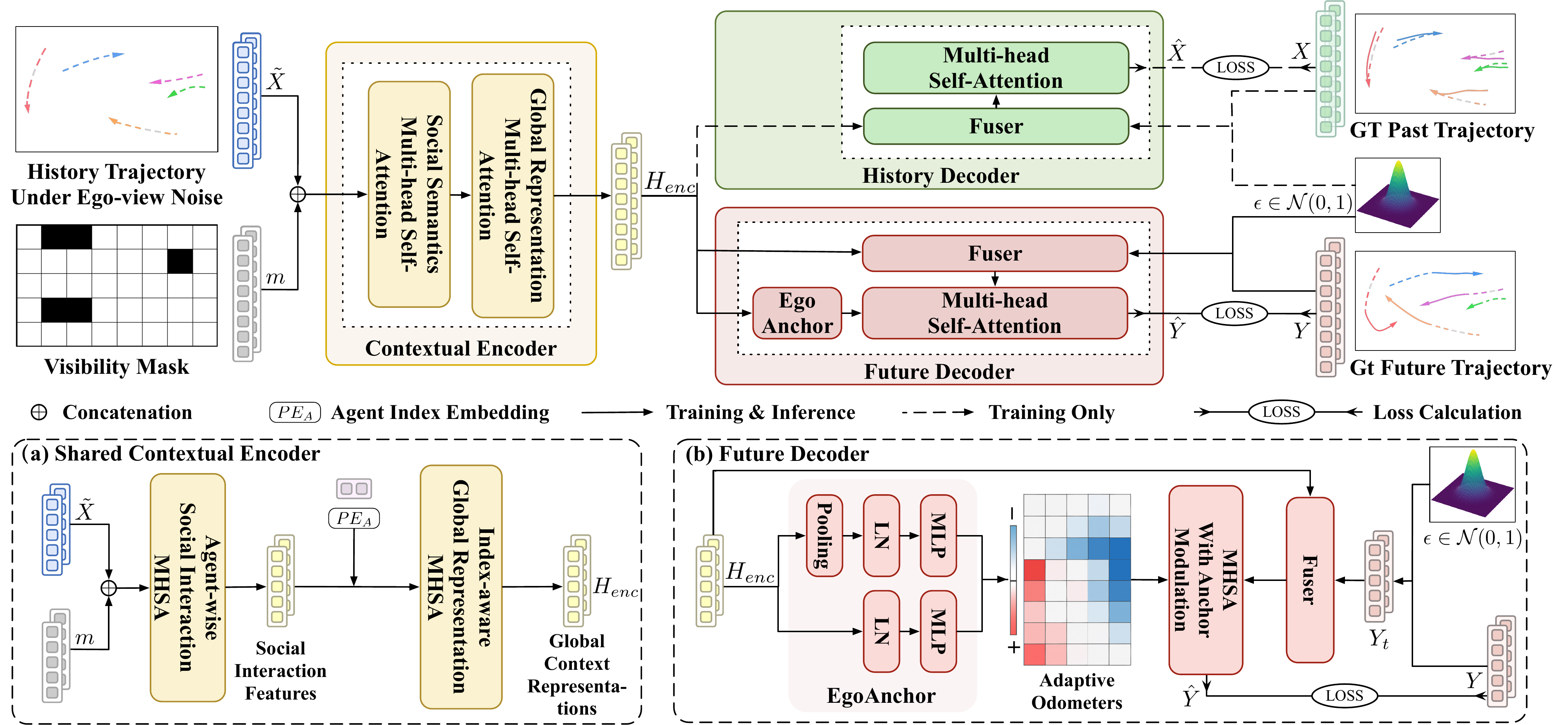}
    \caption{Overview of our BiFlow. The input consists of a noisy historical trajectory $\tilde{X}$ and its corresponding validity mask $m$. During training, the model is supervised with clean ground-truth past $X$ and future $Y$ trajectories to jointly learn reconstruction and prediction. At inference, only the noisy history and mask are used as input to predict the future trajectory $\hat{Y}$.}
    \label{fig:model}
    \vspace{-1em}
\end{figure*}

\subsection{Evaluation Framework and Findings}
\label{sec:eval}

\noindent\textbf{- Datasets.} We utilize EgoTraj-TBD, a real-world pedestrian trajectory dataset with ego-centric perturbation, recorded indoors with complex layouts and dynamic obstacles. In addition, we include the simulated T2FPV dataset for reference, which was collected outdoors across five folds at varying times and locations.

\noindent\textbf{- Metrics.} Following the most commonly used ones in trajectory prediction, we adopt Average Displacement Error (ADE) and Final Displacement Error (FDE) as primary metrics, measuring average and final Euclidean distance between estimated and ground-truth trajectories. Since we evaluate multi-modal methods that generate K candidate trajectories per agent, we report minADE@K and minFDE@K, which compute ADE and FDE of the best-matching trajectory among K outputs, rewarding models for diverse yet accurate predictions. 

\noindent\textbf{- Models.} To assess robustness under noisy FPV observations, we evaluate three classes of state-of-the-art trajectory forecasting models: 1) Recurrent models that iteratively model temporal dynamics through stochastic latent variables, including VRNN\cite{chung2015recurrent}, AC-VRNN\cite{bertugli2021ac}, and SGNet\cite{wang2022stepwise}; 2) Transformer-based models that leverage self-attention mechanisms to capture long-range dependencies and social interactions, represented by TUTR\cite{shi2023trajectory}; and 3) Flow-based generative models that learn data distributions via invertible transformations, specifically MoFlow\cite{fu2025moflow}. This diverse model suite enables a comprehensive analysis of real-world ego-view perceptual challenges. Additionally, to enable a direct comparison with the T2FPV framework, we include its Correction of FPV Errors (CoFE) module\cite{stoler2023t2fpv}, a refinement component trained end-to-end with the prediction model as a baseline for noise correction.

\noindent\textbf{- Empirical Findings.} As shown in Table~\ref{tab:results}, our benchmark reveals a significant gap: all BEV-trained state-of-the-art models exhibit substantial degradation under ego-view perception noise. Particularly, on the widely adopted ETH-UCY datasets, the minADE@20 rises to 0.67m, compared to approximately 0.20m under clean historical trajectories\cite{fu2025moflow}. This highlights a critical limitation that existing methods are highly sensitive to perception artifacts prevalent in ego-view data. These findings underscore the need for noise-aware modeling in trajectory prediction.

\section{Proposed Method}

\subsection{Problem Formulation}

We consider the task of multi-agent trajectory prediction under ego-centric observation noise. 
Let $A$ denote the total number of agents in the scene. 
The input consists of: 
(1) the observed history $T_p$ steps of all agents $\tilde{X} \in \mathbb{R}^{A \times 2T_p}$, 
structured as $\tilde{X} = [\tilde{X}_{\text{other}};\tilde{X}_{\text{ego}}]$, 
where $\tilde{X}_{\text{other}} \in \mathbb{R}^{(A-1) \times 2T_p}$ contains noisy positions of non-ego agents observed from FPV video, 
and $\tilde{X}_{\text{ego}} \in \mathbb{R}^{2T_p}$ contains clean, fully observable positions of the ego-agent from robot odometry; 
(2) a binary validity mask $m \in \{0,1\}^{A \times T_p}$, 
structured as $m = [m_{\text{other}}; \mathbf{1}_{T_p}]$, 
where $m_{\text{other}} \in \{0,1\}^{(A-1) \times T_p}$ indicates validity of non-ego agents, 
and $\mathbf{1}_{T_p}$ denotes full observability of the ego-agent across all $T_p$ steps. 
The goal is to predict the future trajectories $Y \in \mathbb{R}^{A \times 2T_f}$ of all agents over the next $T_f$ time steps.

\subsection{Overview}

Motivated by our benchmark findings that ego-view noise severely degrades the trajectory prediction performance of existing methods, and by the strong performance and efficient conditional generation of recent flow-matching approaches, we propose \textbf{BiFlow}, a dual-stream framework that jointly reconstructs clean agent history trajectories and predicts their future trajectories. 

As illustrated in Fig. \ref{fig:model}, 
BiFlow’s core idea is to transfer denoised motion patterns from history reconstruction to stabilize future prediction under partial or corrupted observations. 
This is realized through three key components: 
(1) a noise-aware contextual encoder that models Social Interactions under occlusion; 
(2) an EgoAnchor mechanism that distills intent priors from history hidden features without future supervision; 
and (3) a dual decoder architecture that shares latent representations of the encoder's output but modulates future prediction via historical confidence. 

Our proposed BiFlow adopts a dual-stream flow matching framework to jointly learn two mappings from the same input $\tilde{X}$: 
(1) reconstructing the clean history trajectory $X_1 \in \mathbb{R}^{A \times 2T_p}$, 
and (2) predicting the clean future trajectory $Y_1 \in \mathbb{R}^{A \times 2T_f}$. 
For each task, we sample a noise vector (i.e., $X_0 \sim {N}(0, I)$ for history, $Y_0 \sim {N}(0, I)$ for future) and construct interpolated states:
\begin{align}
X_t &= (1 - t) X_0 + t X_1, \quad t \in [0,1] \\
Y_t &= (1 - t^{\prime}) Y_0 + t^{\prime} Y_1, \quad t^{\prime} \in [0,1] 
\end{align}

\noindent where $t$ and $t^{\prime}$ are sampled independently to provide diverse supervision in training.

The model is trained to denoise $X_1$ and $Y_1$ conditioned on the observation $\tilde{X}$, time $t$, and interpolated states $X_t$ or $Y_t$, 
using a MoFlow-style\cite{fu2025moflow} multi-candidate objective to ensure trajectory coherence and diversity. 
The total loss combines the flow matching losses from two trajectory prediction tasks,
\begin{equation}
L_{\text{total}} = \lambda_1 L_{\text{recon}} + \lambda_2 L_{\text{pred}},
\end{equation}
where $L_{\text{recon}}$ and $L_{\text{pred}}$ will be elaborated in Sec.~\ref{sec:decoder}.

\subsection{Contextual Encoder: Modeling Noisy Social Dynamics}
\label{sec:encoder}

To capture the noisy dynamics and social interactions of agents’ past trajectories, we design an agent-aware contextual encoder using the Transformer architecture. 

Given the noisy historical trajectory data $\tilde{X}$ and validity masks $m$, we first apply multi-head self-attention (MHSA) to model inter-agent social dependencies. 
We then add a learnable agent-identity encoding $\mathrm{PE}_A$ and apply a second MHSA layer to aggregate a coherent scene-level representation:
\begin{align}
H_{0} &= \mathrm{MLP}\left( \tilde{X} \oplus m \right),\\
H_{\text{soc}} &= \mathrm{MHSA}(Q=H_{0},\, K=H_{0},\, V=H_{0}),\\
H_{\text{soc}}' &= H_{\text{soc}} + \mathrm{PE}_{A},\\
H_{\text{enc}} &= \mathrm{MHSA}(Q=H_{\text{soc}}',\, K=H_{\text{soc}}',\, V=H_{\text{soc}}'),
\end{align}
The output $H_{\text{enc}} \in \mathbb{R}^{A \times D}$ serves as the shared latent representation 
for both reconstruction and prediction streams, where D is the feature dimension.

\begin{table*}[!t]
\centering
\begin{threeparttable}
\caption{Quantitative results across EgoTraj-TBD and T2FPV-ETH using minADE@20 / minFDE@20 (in meters). Best and second-best performances are highlighted in \textbf{bold} and \underline{underlined}, respectively.}
\label{tab:results}
\begin{tabularx}{\textwidth}{@{} l | C C  C C  C C  C C  C C | C @{}}
\toprule
Model $\rightarrow$ 
& \multicolumn{2}{c}{VRNN\cite{chung2015recurrent}}
& \multicolumn{2}{c}{AC-VRNN\cite{bertugli2021ac}}
& \multicolumn{2}{c}{SGNet\cite{wang2022stepwise}}
& \multicolumn{2}{c}{TUTR*\cite{shi2023trajectory}}
& \multicolumn{2}{c|}{MoFlow\cite{fu2025moflow}}
& \multicolumn{1}{c}{\textbf{BiFlow}} \\
\cmidrule(lr){2-3} \cmidrule(lr){4-5} \cmidrule(lr){6-7} \cmidrule(lr){8-9} \cmidrule(lr){10-11} 
Dataset $\downarrow$ & -- & + CoFE & -- & + CoFE & -- & + CoFE & -- & + CoFE & -- & + CoFE & -- \\
\addlinespace[1ex]
\cline{1-12}
\addlinespace[-0.5ex]
\cline{1-12}
\addlinespace[1ex]
ETH$^\dagger$      &   1.35/2.00    &   1.52/2.35    &   1.39/2.04    &   1.47/2.18    &   1.43/1.97    &    0.98/1.32   &    1.25/1.54   &   1.02/1.41    & 0.85/1.22 & 0.88/1.26 & \textbf{0.66/0.85} \\
HOTEL     &    1.30/1.73   &   1.06/1.53    &    1.31/1.75   &   1.16/1.72    &    0.72/1.00   &  0.59/0.76     &   1.04/1.47    &   0.85/1.26   & 0.63/0.89 & 0.62/0.87 &    \textbf{0.49/0.59}     \\
ZARA1     &    1.14/1.68     &  1.65/2.10           &   1.04/1.40    &    1.03/1.48   &   0.58/0.79    &   0.55/0.76    &  0.77/1.00  &  0.62/0.93 & 0.46/0.63 &     0.52/0.67      &   \textbf{0.42/0.58}      \\
ZARA2     &  1.54/2.10  &   1.06/1.63    &   1.47/1.93    &   1.31/1.69    &   0.78/0.91    &    0.73/0.86   &   0.75/0.91   &   0.72/0.82    & \textbf{0.50/0.60} &    0.53/0.63       &      \underline{0.50/0.62}  \\
UNIV      &   2.24/2.89   &  1.27/1.62     &  2.26/3.00  &   1.54/1.88    &   1.48/1.73    &  1.23/1.48     &  1.10/1.30   &    1.02/1.24   &     \underline{0.92/1.10}     &  0.92/1.08         &     \textbf{0.91/1.08}    \\
\addlinespace[0ex]
\midrule
\addlinespace[0.5ex]
AVG &  1.51/2.08 &   1.31/1.84    &   1.49/2.03    &  1.30/1.79     &    1.00/1.28   &   0.82/1.04    &   0.98/1.12    &    0.85/1.13    &    0.67/0.88       &     0.69/0.90     &     \textbf{0.60/0.74}    \\
\addlinespace[0.6ex]
\cline{1-12}
\addlinespace[-0.5ex]
\cline{1-12}
\addlinespace[1.3ex]
\textbf{Ego-TBD}       &   0.76/1.26    &   0.68/1.19    &    0.82/1.38   &   0.63/1.09    &   0.34/0.52    &    0.37/0.58   &   0.58/0.72    &  0.52/0.67   &      \underline{0.21/0.29}     &     0.26/0.36      &   \textbf{0.19/0.27}      \\
\bottomrule
\end{tabularx}

\begin{tablenotes}
\small
\item[$\dagger$] ETH/HOTEL/ZARA1/ZARA2/UNIV are the five folds from T2FPV-ETH.
\item[*] The TUTR architecture is adapted to support multi-modal output to ensure a fair comparison.
\end{tablenotes}
\end{threeparttable}
\vspace{-1em}
\end{table*}

\subsection{EgoAnchor: Intent Prior Distillation}
\label{sec:egoanchor}
Drawing on intent-driven models\cite{mangalam2021goals, wang2022stepwise} that infer long-term
goals to guide behavior prediction, we introduce \textbf{EgoAnchor}, a lightweight mechanism that distills intent priors from $H_{\text{enc}}$. These priors are constructed to encode historical context to stabilize predictions under partial or corrupted observations. 
\begin{align}
Ach_{\text{agent}} &= \mathrm{MLP}(\mathrm{LayerNorm}(H_{\text{enc}})), \\
Ach_{\text{scene}} &= \mathrm{MLP}(\mathrm{LayerNorm}(\mathrm{Mean}(Ach_{\text{agent}}))),
\end{align}
where $Ach_{\text{agent}} \in \mathbb{R}^{A \times D}$ is used to capture agent-level motion tendencies, 
and $Ach_{\text{scene}} \in \mathbb{R}^{D}$ summarizes global context. We use layer normalization to handle crowd size, validity, and camera noise variations in $H_{enc}$, making the output anchor comparable across agents and scenes. 

To reduce computation, EgoAnchor operates in a self-supervised manner by eschewing additional intent labels. The extracted anchors are integrated into the future decoder via feature-wise affine modulation~\cite{perez2018film}, modulating feature distribution based on agent/scene-level intent and reliability.

\subsection{Dual Decoder with Multi-Candidate Prediction}
\label{sec:decoder}

Motivated by leveraging clean motion patterns from history reconstruction to future prediction, we employ two independent decoders that share the encoder output $H_{enc}$, as shown in Figure~\ref{fig:model}.

In \textbf{future prediction} stream, we predict clean future trajectory $Y_1$ from $H_{enc}$, time step $t^{\prime}$, and noisy intermediate state $Y_{t^{\prime}}$. These inputs are first fused into a hidden state, which is then processed through $K$-to-$K$ MHSA blocks to model interactions among $K$ candidate trajectories for diverse yet coherent predictions. The resulting hidden state $Z$ is modulated using intent priors from the EgoAnchor module:
\begin{align}
\bigl(\beta_{Ach},\gamma_{Ach}\bigr)&=\mathrm{MLP}(Ach_{\mathrm{agent}}+Ach_{\mathrm{scene}}), \\
Z_{Ach} &= (1+\gamma_{Ach}) \,\odot\, Z + \beta_{Ach}, \\
H_{dec} &= \mathrm{MHSA} (Q=Z_{Ach},K=Z_{Ach},V=Z_{Ach})
\end{align}
\noindent where $\beta_{Ach}$ and $\gamma_{Ach}$ are affine parameters adaptively modulating feature space based on historical confidence, analogous to adaptive odometers (e.g, amplifying high-confidence features). To refine inter-agent dependencies, a subsequent MHSA is employed. Finally, an MLP head maps $H_{dec}$ to candidate predictions $\hat{Y}^{1:K}$ and logits $\hat{c_y}$. The training objective for the future prediction task is: 
\begin{align}
L_{pred} &= \mathbb{E}_{t^{\prime},\,Y_{1},\,Y_{t}}\!\left[\bigl\|\hat{Y}^{1:K}-Y_{1}\bigr\|_{2}^{2}+\mathrm{CE}(c_{y}^{1:K},j^{*}_{y})\right]  \\
j^{*}_y &= \arg\min_{j} \|\hat{Y}^j - Y_1\|_2^2
\end{align}
\noindent where $j^*$ identifies the best-matching candidate, and $\mathrm{CE}(\cdot,\cdot)$ is the cross-entropy loss over mode selection. 

The \textbf{history reconstruction} stream mirrors identical structure but sets $\beta_{Ach}$ and $\gamma_{Ach}$ to 0, enabling pure reconstruction learning. 
During inference, the history reconstruction decoder is no longer activated. Instead, we use only the noisy input trajectory $\tilde{X}$, which is processed through the shared contextual encoder and the prediction decoder and conditioned on intent priors from the EgoAnchor module, to forecast future trajectories.

\section{Experiments}

\subsection{Implementation Details}

We evaluate BiFlow and existing baseline methods on our proposed EgoTraj-Bench, as detailed in Sec.~\ref{sec:eval}. In our approach, all trajectories are normalized to stabilize training. In the history reconstruction branch, we use absolute coordinates as targets to facilitate denoising; in the future prediction branch, we adopt displacement-based (relative) targets to improve temporal coherence. For feature modulation in the future stream, we integrate EgoAnchor through a 4-layer MHSA block, which enables progressive fusion of structured prior information into the decoder. We sample from the model using 10 denoising steps based on a logit-normal time scheduler. The model is trained for 150 epochs with a batch size of 64 and a latent dimension of 128. We use the AdamW optimizer with an initial learning rate of 0.001, a cosine annealing learning rate schedule with warmup, and a weight decay of 0.01.

\begin{table}[!t]
\centering
\caption{Ablation study on EgoTraj-TBD.}
\label{tab:ablation}
\setlength{\tabcolsep}{8pt}        
\renewcommand{\arraystretch}{1.0}  

\begin{tabular}{@{} 
    l 
    | 
    c c c
    | 
    c c c 
    @{}}
\toprule
Model & \multicolumn{3}{c|}{Components} & \multicolumn{3}{c}{ADE/FDE@K} \\
\cmidrule(lr){2-4} \cmidrule(lr){5-7}
& SI & EA & SE & k=1 & k=5 & k=10 \\
\addlinespace[1ex]
\cline{1-7} 
\addlinespace[-0.5ex] 
\cline{1-7} 
\addlinespace[1ex]
MoFlow & - & - & - & 0.84/1.45 & 0.46/0.76 & 0.31/0.48 \\
\addlinespace[1ex]
\cline{1-7} 
\addlinespace[-0.5ex] 
\cline{1-7} 
\addlinespace[1ex]
\multirow{3}{*}{\textbf{BiFlow}} & \ding{52} & \ding{55} & \ding{55} & 0.76/1.37 & 0.42/0.72 & 0.29/0.47 \\
& \ding{52} & \ding{52} & \ding{55}& 0.73/1.32& 0.40/0.69& 0.28/0.45\\
& \ding{52} & \ding{52} & \ding{52} & \textbf{0.70/1.21}& \textbf{0.38/0.63}& \textbf{0.26/0.41} \\
\bottomrule
\end{tabular}
\end{table}

\subsection{Quantitative Results}

We present performance comparisons across all models on both datasets in Table~\ref{tab:results}. 
Across architectures, ego-view noise substantially degrades performance relative to clean-history results.
Recurrent models (VRNN/AC-VRNN/SGNet) degrade more severely due to error accumulation under corrupted histories, while the attention-based predictor TUTR is generally more competitive but still struggles under realistic noise. Among prior methods, the flow-matching model MoFlow is the strongest competitor, suggesting that conditional generative modeling is promising for forecasting from noisy histories, yet it still degrades under corrupted observations. 

Under ego-view noise, our method is both more robust and more efficient. On T2FPV-ETH, it achieves state-of-the-art (SOTA) performance with minADE@20 of 0.60 and minFDE@20 of 0.74, outperforming the prior SOTA by over 11\% and 15\%, respectively.
On EgoTraj-TBD, it shows consistent improvements over existing methods. Notably, as shown in Table~\ref{tab:ablation}, our model achieves significant improvements when generating fewer future trajectory candidates (i.e., smaller $K$). Specifically, with the same reduced number of samples, our approach improves minADE and minFDE by around 16\% compared to the prior SOTA. The strong performance indicates a predicted distribution more closely aligned with true trajectories, underscoring enhanced robustness and predictive efficiency under noisy conditions.

Additionally, while integrating CoFE brings improvements for some models, the gains are limited, 
and flow-based methods can even degrade.
This suggests that correcting only missing positions is insufficient, as ego-centric observations contain structured noise beyond occlusion, including tracking errors and perspective distortion. Effective denoising therefore requires holistic trajectory modeling rather than patchwise correction, which is validated by our SOTA results.

Note that for all models, including BiFlow, the ADE/FDE values on EgoTraj-TBD are lower than those evaluated on T2FPV-ETH. A key reason is our higher-fidelity ground truth and more noise-aware processing pipeline, which results in significantly a lower historical MSE (0.66m) in EgoTraj-TBD, as shown in Table~\ref{tab:benchmark}. These more meaningful historical representations enable more reliable training and contribute to overall better performance across methods. 

\subsection{Ablation Study}

Table~\ref{tab:ablation} presents an ablation study of BiFlow on EgoTraj-TBD: Social Interaction (SI) within the contextual encoder, EgoAnchor (EA) distillation, and Shared Encoder (SE) used by both streams. 
We compare against MoFlow as a strong baseline, which can be viewed as a representative single-flow predictor that directly forecasts futures from past motion.
Although it is not a strictly architecture-matched dual-versus-single control, BiFlow’s consistent improvements over MoFlow suggest that jointly modeling history denoising and future forecasting is beneficial under ego-view noise.

We then ablate components by enabling them progressively. When only SI is added, BiFlow achieves notable improvements over MoFlow, reducing minADE/minFDE by over 9\% at $K$=1. Incorporating EA further enhances performance, yielding a 13\% improvement with the same $K$. With all components (SI, EA, and SE) integrated, the full model achieves the best performance, improving minADE and minFDE by 16\%, 17\%, and 16\% at $K$=1, $K$=5, and $K$=10, respectively. These results demonstrate that integrating social interaction cues and ego-centric intent patterns, particularly through EgoAnchor, significantly improves prediction accuracy and robustness under the deployment-realistic condition.

\begin{figure}[H]
    \centering
    \makebox[\linewidth][c]{\includegraphics[width=1\linewidth,trim={0.cm 0cm 0cm 0cm}, clip]{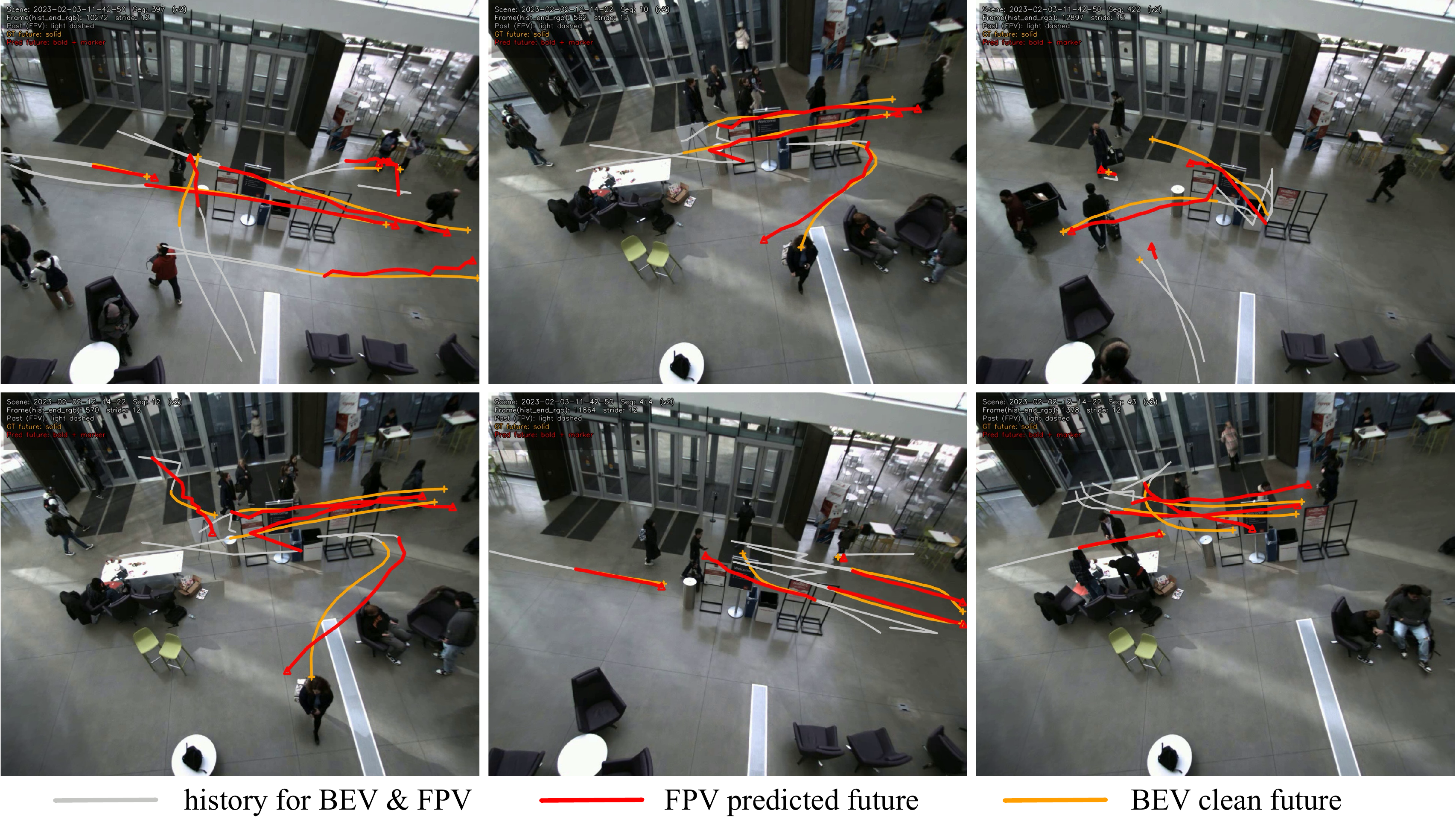}}
    \caption{Visualization under ego-view noise. \textbf{Gray}: past trajectories (BEV and noisy FPV-derived histories). \textbf{Orange}: clean BEV futures (GT). \textbf{Red}: predicted futures.}
    \label{fig:qualitative}
    \vspace{-0em}
\end{figure}

\subsection{Qualitative Results}


Fig.~\ref{fig:qualitative} provides qualitative examples on EgoTraj-TBD. Although the BEV history and the FPV-derived history (both shown in gray) can differ substantially due to ego-view perception noise, BiFlow predicts future trajectories (red) that closely follow the clean BEV ground truth futures (orange). The examples highlight that BiFlow maintains accurate and physically plausible motion under deployment-realistic history corruption, indicating improved robustness to noisy and incomplete observations.

\section{Conclusion}

We introduce EgoTraj-Bench, a new benchmark that pairs noisy first-person-view observations with human-verified metric-space ground truth, capturing authentic deployment-level perturbations. The benchmark highlights a critical gap in trajectory prediction: the disconnect between idealized BEV-based evaluation and real-world ego-centric perception noise. We further propose BiFlow, a dual-stream framework with the EgoAnchor mechanism for intent-aware prediction. Experiments on our EgoTraj-Bench verify the effectiveness of our designs and demonstrate clear advantages in noisy and resource-constrained settings. 

While our approach shows competitive performance in the current setup, our benchmark relies on projecting FPV-derived tracks into metric space, which may introduce residual alignment error, and we do not yet quantify the impact of individual noise types.
Generalization may also vary with camera geometry and sensing characteristics. 
Future work will improve calibration-robust projection, noise-stratified evaluation, and adaptation to diverse platforms and environments.
We hope that EgoTraj-Bench and BiFlow will support the community with models more robust to the ego-centric observation challenges and progress toward reliable real-world deployment.



\addtolength{\textheight}{-1cm}   








\bibliographystyle{IEEEtran}
\bibliography{bibtex/bib/IEEEabrv, main}

\end{document}